%% file: TDPTemplate.tex
\documentclass[runningheads,a4paper]{llncs}
\usepackage{amssymb}
\usepackage{graphicx}
\usepackage{amssymb}
\usepackage[utf8]{inputenc}
\usepackage{url}
\usepackage{float}
\usepackage{amsmath}
\usepackage{graphicx}
\usepackage{wrapfig}
\usepackage{booktabs}
\usepackage{multirow}
\usepackage{url}
\usepackage{lipsum}

\begin{document}

\title{Hibikino-Musashi@Home\\2018 Team Description Paper}

\author{Yutaro Ishida \and Sansei Hori \and Yuichiro Tanaka \and Yuma Yoshimoto \and \\ Kouhei Hashimoto \and Gouki Iwamoto \and Yoshiya Aratani \and Kenya Yamashita \and \\ Shinya Ishimoto \and Kyosuke Hitaka \and Fumiaki Yamaguchi \and \\ Ryuhei Miyoshi \and Kentaro Honda \and Yushi Abe \and Yoshitaka Kato \and \\ Takashi Morie \and Hakaru Tamukoh }
\institute{Graduate school of life science and systems engineering,\\Kyushu Institute of Technology, \\
2-4 Hibikino, Wakamatsu-ku, Kitakyushu 808-0196, Japan, \\
\texttt{http://www.brain.kyutech.ac.jp/\~{}hma/wordpress/}}
\authorrunning{Yutaro Ishida et al.}
\maketitle


\begin{abstract}
Our team, Hibikino-Musashi@Home (the shortened name is HMA), was founded in 2010. It is based in the Kitakyushu Science and Research Park, Japan.
We have participated in the RoboCup@Home Japan open competition open platform league every year since 2010.
Moreover, we participated in the RoboCup 2017 Nagoya as open platform league and domestic standard platform league teams.
Currently, the Hibikino-Musashi@Home team has 20 members from seven different laboratories based in the Kyushu Institute of Technology.
In this paper, we introduce the activities of our team and the technologies.

\end{abstract}


\section{Introduction}
Our team, Hibikino-Musashi@Home (the shortened name is HMA), was founded in 2010 and it competes in the RoboCup@Home Japan Open open platform league (OPL) every year.
Our team is based in the Kitakyushu Science and Research Park and we have 20 team members from seven different laboratories of the Kyushu Institute of Technology.
We are currently developing a home-service robot and we intend to use this event to present the outcomes of our research.\par

In the Intelligent Home Robotics (iHR) challenge, which is a competition that takes place in Japan, we were awarded the first prize.
This competition included a manipulation and object recognition test as well as a speech-recognition and audio-detection test; these tests are the same as those generally used in the RoboCup@Home competitions.
In the RoboCup 2017 Nagoya, we participated in the OPL and domestic standard platform league (DSPL).
Furthermore, we were awarded the first prize in the DSPL using a TOYOTA HSR robot \cite{toyota_hsr}.


This paper describes the technologies that we used in the iHR challenge and the RoboCup 2017 Nagoya.
In particular, this paper outlines both an object recognition system that uses deep learning \cite{hinton2006fast}, a speech recognition system and a sound localization system installed in our HSR.

\section{System overview}
Hibikino-Musashi@Home has used an HSR since 2016.
We were able to customize the HSR in the iHR challenge.
Therefore, we used a customized HSR similar to a rule ``Official Standard Laptop for DSPL'' of RoboCup@Home 2018.
In addition, we use a non-customized HSR in the RoboCup 2017 Nagoya.
In this section, we introduce the specifications of our customized and non-customized HSRs.

\subsection{Customized HSR}
\subsubsection{Hardware overview}
\begin{figure}[t]
\begin{center}
\includegraphics[width=7cm]{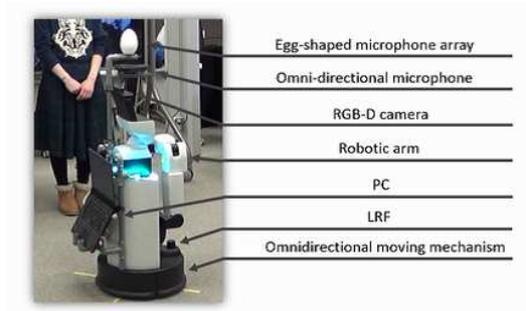}
\caption{Devices on the customized HSR.}
\label{fig:cust_hsr}
\end{center}
\end{figure}

Hibikino-Musashi@Home uses a customized HSR shown in Fig. \ref{fig:cust_hsr}.
We participated in the iHR 2016 challenge with it.
The computational resources built into the HSR were not sufficient to support our intelligent systems and extract the maximum performance from the system.
To solve this problem, we developed and incorporated our own customizations.
As part of our customization, we added a laptop computer that could handle the computational requirements of our intelligent systems.
We also added to the HSR an omnidirectional microphone and an egg-shaped microphone array that would facilitate sound localization so that speech could be recognized.
Table \ref{tab:extDev} shows specification of the devices attached to our robot.
Consequently, the computer inside the HSR could be used solely to run the HSR's basic software such as its airframe control.
This enabled the HSR to operate with more stability.
Further, by incorporating additional microphones, we were able to facilitate ambient sound recognition for our HSR.

\begin{table}[t]
\begin{center}
\caption{Specification of the devices on the customized HSR.}
\label{tab:extDev}
\begin{tabular}{cc}\toprule
	Computer		&		ThinkPad PC Core-i5 4850U processor and 12Gb RAM x2 \\
	Omnidirectional mic.	&		YAMAHA PJP-20UR \\
	Egg-shaped mic. array	&		TAMAGO-03 \\
\bottomrule
\end{tabular}
\end{center}
\end{table}

\subsubsection{Software overview}
\begin{figure}[tb]
\begin{center}
\includegraphics[scale=0.5]{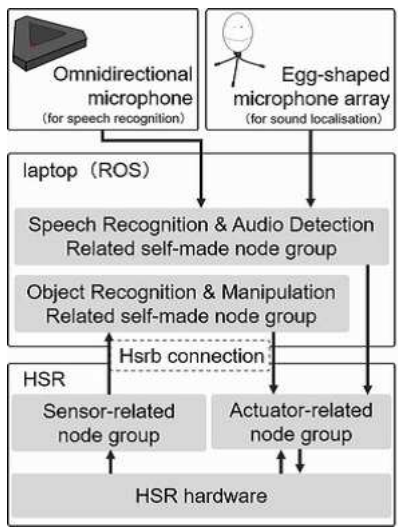}
\caption{Block diagram for an overview of the customized HSR software blocks.}
\label{fig:softOverview1}
\end{center}
\end{figure}

In this section, we introduce the software that we installed in our HSR for the iHR challenge.
We tackled the manipulation and object recognition test and the speech and audio recognition test by using an image recognition system that utilizes deep learning and a voice recognition system, respectively.
Figure \ref{fig:softOverview1} shows the customized system that we installed in our HSR.
The system is based on Robot Operating System (ROS)\cite{ros}, which is the same infrastructure that we have used for our OPL robot.
As such, we have been able to develop a system for our HSR rapidly by porting over the software developed for our OPL robot, including our ``state management'', ``voice interaction'', and ``image processing'' algorithms.
In our customized HSR system, the laptop computer is used to process the system.
This laptop was connected to the computer embedded in the HSR through an Hsrb interface.
The built-in computer specializes in low-layer systems, such as the HSR's airframe control system and the sensor system.

\begin{figure}[tb]
\begin{center}
\includegraphics[scale=0.55]{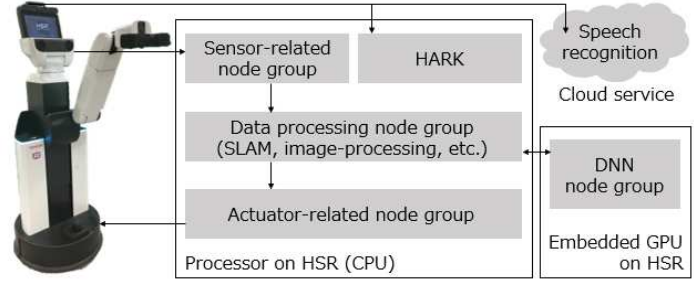}
\caption{Block diagram for an overview of the non-customized HSR software blocks.}
\label{fig:softOverview2}
\end{center}
\end{figure}

\subsection{Non-customized HSR}
In this section, we introduce a system that we installed in a non-customized HSR for the RoboCup 2017 Nagoya (Fig. \ref{fig:softOverview2}).
The difference between the customized HSR and the non-customized HSR is whether the data processing node group is processed by the external laptop PC or a processor on HSR.


In the system, an embedded GPU is used to process the system.
This embedded GPU is connected to an processor in the HSR through an Ethernet connection.
The embedded GPU specializes to accelerate the infarance of deep neural networks (DNNs).


\section{Object recognition}
In this section, we explain the object recognition system that employs deep learning, which we used in the RoboCup 2017 Nagoya.

\subsection{Processing flow of the object recognition system}
\begin{figure}[tb]
\begin{minipage}{0.5\hsize}
	\begin{center}
		\includegraphics[height=5cm]{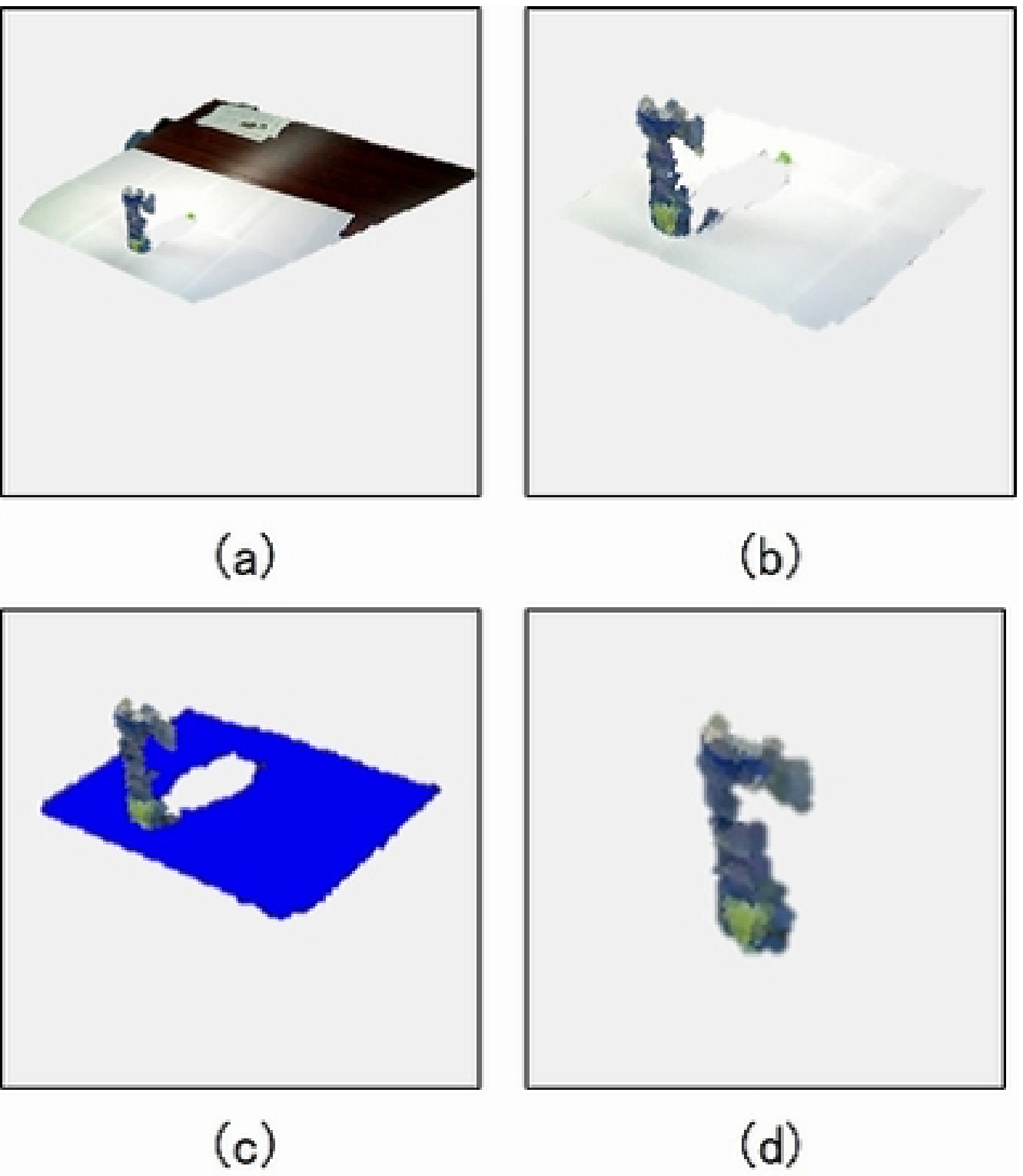}
		\caption{Processing flow of the object\newline recognition system.}
		\label{fig:imgRecogFlow}
	\end{center}
\end{minipage}
\begin{minipage}{0.5\hsize}
	\begin{center}
		\includegraphics[height=5cm]{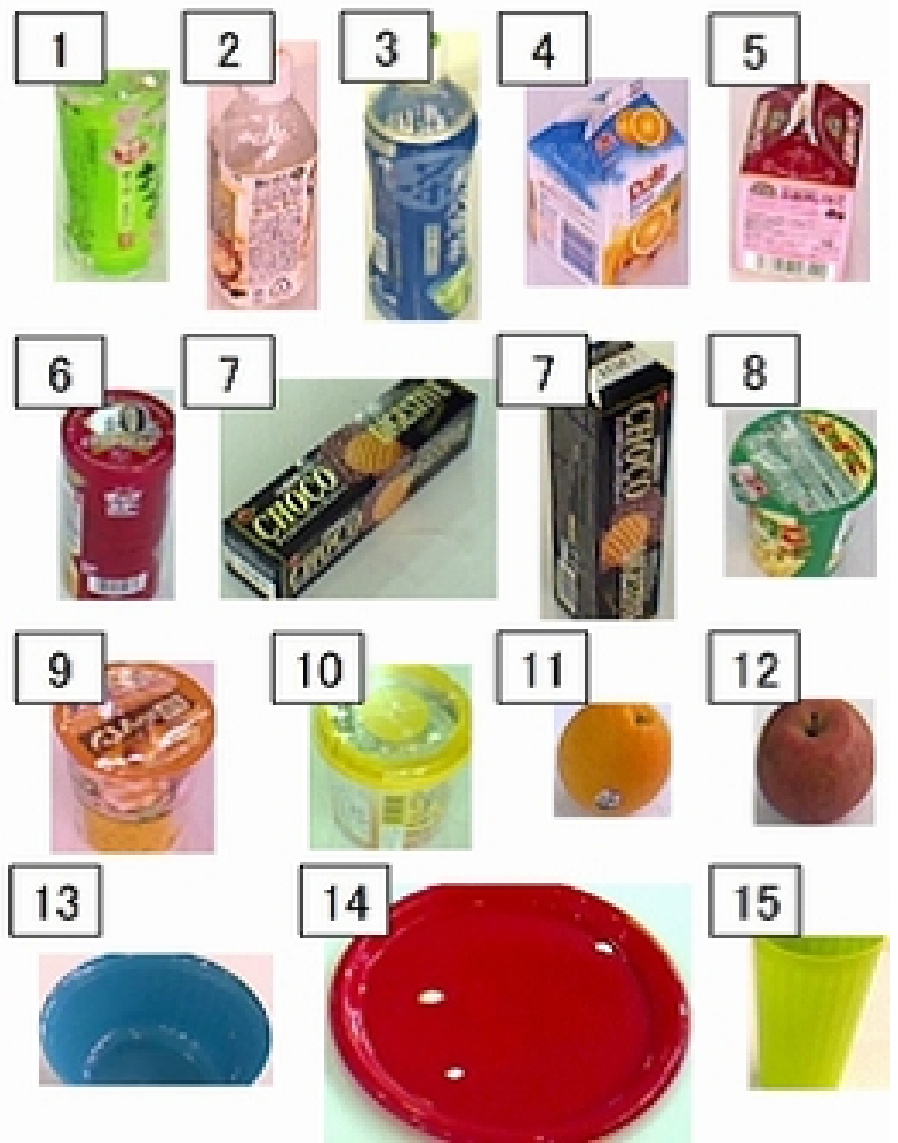}
		\caption{Objects used for the RoboCup Japan Open 2016.}
		\label{fig:objects}
	\end{center}
\end{minipage}
\end{figure}

Our object recognition system has two major parts, which are an object extraction system that obtains images from a camera, and an object recognition system.
The entire processing flow in the object extraction and object recognition systems are as follows:

\begin{enumerate}
\item Obtain a target 3D point cloud as shown in Fig. \ref{fig:imgRecogFlow}(b) by applying a pass through-filter to an original 3D point cloud as shown in Fig. \ref{fig:imgRecogFlow}(a) using a point cloud library (PCL). The image is obtained using an RGB-D camera on the HSR.

\item Apply a random sample consensus method, which includes the PCL, to the target 3D point cloud; this is used to detect the tabletop shown using blue color in Fig. \ref{fig:imgRecogFlow}(c).

\item Clip a 3D point cloud for the object, which is on the table, as shown in Fig. \ref{fig:imgRecogFlow}(d), from the target 3D point cloud. 

\item Clip an object image from the RGB image using the object 3D point cloud in order to obtain an object image.

\item Classify the object using a DNN.

\item Calculate the object coordinates using the RGB-D camera coordinate system and the object 3D point cloud.

\item Change the object coordinates from the RGB-D camera coordinate system to the robot coordinate system.

\item Control a manipulator so that it moves toward the target coordinates and grasps the object.
\end{enumerate}

\subsection{Classification methods}
We use Caffe \cite{caffe}, which is a deep learning framework, and OpenCV \cite{opencv}, which is an image processing library, to construct the object recognition system.
For the deep learning system, we use GoogLeNet \cite{szegedy2015going}, which is a 22-layered convolutional neural network that learns using the dataset provided by the Image Net Large-Scale Visual Recognition Challenge 2012 \cite{imagenet}. We employ transfer learning to GoogLeNet, and the final layer is fine-tuned using the dataset that we have created for RoboCup@Home. We used a training dataset that has images that are 224 $\times$ 224 pixels in size, RGB-3Ch, and have 8-bit gradation.

\subsection{Experiments and results}
Our object recognition system was trained using images of 15 objects as shown in Fig. \ref{fig:objects}. These objects were used in the RoboCup Japan Open 2016 and they were used to evaluate the robots that competed. Images of these objects were captured from various angles while being rotated in order to create a dataset. After this, each pixel value of each image was multiplied by 0.9, 1.0, and 1.1 in order to add noise. Consequently, our dataset contained 2,700 images of each object. We used 2,000 of these images as the training data for the HSR.

Table \ref{tab:result} shows the results of our evaluation of the object recognition system with the test dataset which included 12 objects numbered one to twelve as shown in Fig. \ref{fig:objects}. The test dataset was captured from 12 different directions at 30 degree increments in an environment that was different from the environment where the images for the training dataset were taken; this was done so that we could evaluate how the system responded to environmental changes. From the experimental results, we determined that our object recognition system performed to a high degree of accuracy for the RoboCup@Home tasks.

\begin{table}[tb]
\begin{center}
\caption{Classification results with GoogLeNet.}
\label{tab:result}
\begin{tabular}{|c|c|c|c|c|c|c|c|c|c|c|c|c|c|c|}\hline
	\multirow{2}{*}{Method} & & \multicolumn{12}{c|}{Object number} & \multirow{2}{*}{Accuracy rate [\%]} \\ \cline{3-14}
	 & & 1 & 2 & 3 & 4 & 5 & 6 & 7 & 8 & 9 & 10 & 11 & 12 & \\ \hline \hline
	\multirow{2}{*}{GoogLeNet2014} & True & 12 & 12 & 12 & 11 & 12 & 12 & 12 & 12 & 12 & 12 & 12 & 12 & \multirow{2}{*}{99} \\ \cline{2-14}
	 & False & 0 & 0 & 0 & 1 & 0 & 0 & 0 & 0 & 0 & 0 & 0 & 0 &  \\ \hline
\end{tabular}
\end{center}
\end{table}


\section{Speech recognition and sound localization}
The voice recognition/sound source localization system operates as follows:

\begin{enumerate}
\item The microphone array captures the voice of the person addressing it and uses a speech recognition engine, Web Speech API on google chrome, to recognize that it is being spoken to.

\item The voice of the person addressing it is captured by the microphone array, and a sound source localization is performed by the MUSIC method using HARK \cite{hark}, which is a piece of auditory software for robots.
\end{enumerate}

\section{Result of the RoboCup 2017 Nagoya}
We participated in the RoboCup 2017 Nagoya using the described system.
We scored 524 points out of 2125.
This points were 87.2\% of the top OPL team's points.
We were able to show the performance of new robot HSR and our technoloies.
Thanks to these results, we were awarded the first prize in the competition.

\vspace{-0.3cm}
\section{Conclusions}
In this paper, we have summarized available information about our HSR that participated in the iHR challenge and the RoboCup 2017 Nagoya.
The object recognition and voice interaction capabilities that we built into the robot are also described.
Currently, we are developing a number of different pieces of software for HSRs to participate in the RoboCup 2018 Montreal.

\vspace{-0.3cm}
\section*{GitHub}
Source codes of our systems and our original dataset are published on GitHub. The URL is as follows:\\
https://github.com/hibikino-musashi-athome

\vspace{-0.3cm}
\section*{Acknowledgement}
This work was supported by Ministry of Education, Culture, Sports, Science and Technology, Joint Graduate School Intelligent Car \& Robotics course (2012-2017), Kitakyushu Foundation for the Advancement of Industry Science and Technology (2013-2015), Kyushu Institute of Technology 100th anniversary commemoration project : student project (2015) and YASKAWA electric corporation project (2016-2017), JSPS KAKENHI grant number 17H01798, and the New Energy and Industrial Technology Development Organization (NEDO).

\vspace{-0.3cm}
\bibliography{ref.bib}

\newpage
\input{RobotDescription}

\end{document}

%% file: RobotDescription.tex
\section*{Robot HSR Hardware Description}
In this section briefly describe the hardware of the robot

\begin{itemize}
	\item Name: Human support robot (HSR).
	\item Footprint: 430 mm.
	\item Height(min/max): 1005/1350 (top of the head height).
	\item Weight: About 37 kg.
	\item Sensors on the moving base:
		\begin{itemize}
			\item Laser range sensor.
			\item IMU.
		\end{itemize}
	\item Arm length: 600 mm.
	\item Arm payload (recommended/max): 0.5/1.2 kg.
	\item Sensors on the gripper:
		\begin{itemize}
			\item Gripping force sensor.
			\item Wide-angle camera.
		\end{itemize}
	\item Sensors on the head:
		\begin{itemize}
			\item RGB-D sensor.
			\item Stereo camera.
			\item Wide-angle camera.
			\item Microphone array.
		\end{itemize}
	\item Body expandability: USB x3, VGA x1, LAN x1, Serial x1, and 15V-0.5A output x1.
\end{itemize}


\section*{Robot's Software Description}
For our robot we are using the following software:

\begin{itemize}
	\item OS: Ubuntu 14.04.
	\item Middleware: ROS Indigo.
	\item State management: SMACH (ROS).
	\item Speech recognition (English):
		\begin{itemize}
			\item rospeex\cite{rospeex}.
			\item Web Speech API.
			\item IBM Watson Speech To Text.
		\end{itemize}
	\item Morphological Analysis Dependency Structure Analysis (English): SyntaxNet.
	\item Speech synthesis (English): Web Speech API.
	\item Speech recognition (Japanese): Julius.
	\item Morphological Analysis (Japanese): MeCab.
	\item Dependency structure analysis (Japanese): CaboCha.
	\item Speech synthesis (Japanese): Open JTalk.
	\item Sound location: HARK.
	\item Object detection: Point cloud library (PCL) and You only look once (YOLO)\cite{redmon2016you}.
	\item Object recognition: Caffe with GoogLeNet and YOLO.
	\item Human detection / tracking:
		\begin{itemize}
			\item Depth image + particle filter.
			\item OpenPose\cite{cao2017realtime}.
		\end{itemize}
	\item Face detection: SkyBiometory.
	\item SLAM: slam\_gmapping (ROS).
	\item Path planning: move\_base (ROS).
\end{itemize}